\documentclass{article}
\usepackage{tikz}
\usetikzlibrary{arrows.meta, positioning} 
\usetikzlibrary{fit}
\usepackage{natbib}
\usepackage[utf8]{inputenc} 
\usepackage[T1]{fontenc}    
\usepackage{hyperref}       
\usepackage{url}            
\usepackage{booktabs}       
\usepackage{amsfonts}       
\usepackage{nicefrac}       
\usepackage{microtype}      
\usepackage{xcolor}         

\usepackage{outlines}
\usepackage{amssymb}
\usepackage{bm}
\usepackage{bbm}
\usepackage{xcolor}
\usepackage{enumerate}
\usepackage{ulem}
\usepackage{parskip}
\usepackage{graphicx}
\usepackage{subcaption}

\usepackage{algpseudocode}
\newcommand{\bs}[1]{\boldsymbol{#1}}
\usepackage{amsmath}
\usepackage{algorithm} 
\usepackage{tikz}
\usepackage{cleveref}

\title{A Hierarchical Variational Graph Fused Lasso for Recovering Relative Rates in Spatial Compositional Data}

\author{%
  Joaquim Valerio Teixeira, Ed Reznik, Sudpito Banerjee, Wesley Tansey
}

\begin{document}

\maketitle

\begin{abstract}
    The analysis of spatial data from biological imaging technology, such as imaging mass spectrometry (IMS) or imaging mass cytometry (IMC), is challenging because of a competitive sampling process which convolves signals from molecules in a single pixel. To address this, we develop a scalable Bayesian framework that leverages natural sparsity in spatial signal patterns to recover relative rates for each molecule across the entire image. Our method relies on the use of a heavy-tailed variant of the graphical lasso prior and a novel hierarchical variational family, enabling efficient inference via automatic differentiation variational inference. Simulation results show that our approach outperforms state-of-the-practice point estimate methodologies in IMS, and has superior posterior coverage than mean-field variational inference techniques. Results on real IMS data demonstrate that our approach better recovers the true anatomical structure of known tissue, removes artifacts, and detects active regions missed by the standard analysis approach.
\end{abstract}

\pagenumbering{arabic}

\pagenumbering{arabic}
\section{Introduction}
The last five years have seen an explosion in the prevalence and prominence of spatial profiling technologies, such as spatial transcriptomics~\citep{marx:2021:st-method-of-the-year} and spatial proteomics~\citep{karimi:2024:spatial-proteomics-method-of-the-year}, in biological research. These technologies enable characterization of the abundance of molecules across a spatial structure, such as tissue, cell, or organ samples. Analysis of spatial profiling data is often limited by the competitive sampling nature of many profiling technologies, which convolve signals from co-localized molecules. Although scientists are interested in analyzing relative intensity rates across pixels within a single molecular type, the data produced by these technologies provides within-pixel relative rates across molecular types.~\looseness=-1

The fundamental problem with conflating within-pixel rates with within-molecule rates is that it is impossible to identify the latter from the former without prior knowledge or limiting assumptions~\cite{pawlowsky:etal:2015:modeling-compositional-data}. Consider the case where one pixel reports a composition of (0.1, 0.4, 0.5) for molecules A, B, and C, respectively, and another reports (0.15, 0.6, 0.25). The compositional system is underdetermined, making it impossible to know from the compositional readout whether the second pixel saw a decrease in molecule C or an increase in both molecules A and B, or a mixture of both. Simply comparing proportions naively can lead to incorrect interpretation of the data.~\looseness=-1

To address this issue, we have developed a statistical learning methodology that leverages the biological knowledge that molecular abundances relate to the underlying structure of cells in a tissue. Cells tend to organize into spatially contiguous tissue subregions~\cite{bryant:mostov:2008:cells-organize}, leading to a piecewise-constant pattern of molecule abundances across a spatial domain. We show, perhaps surprisingly, that encoding this biological knowledge in the form of a sparse hierarchical graphical model is sufficient to recover the true relative rates of individual molecules across pixels. By imposing a data-dependent sparsity between the change in rates of the same molecule in neighboring pixels, our approach reduces the degrees of freedom of the overall system, empirically enabling identification of the true rates.~\looseness=-1



To scale inference to large datasets generated by modern spatial profiling technologies, we develop a novel structured variational inference (VI) algorithm. We use a hierarchical variational distribution over the latent log-rates and the edge-specific shrinkage priors to allow for latent rates at points of change to have higher variances. Unlike the standard mean-field VI, our approach imparts an implied spatial dependence on the joint variational posterior distribution of the logarithmic rates, producing well-calibrated joint posteriors while allowing for the efficiency of conditionally independent sampling in the gradient calculation.~\looseness=-1

\begin{figure}[t!]
    \centering
    \includegraphics[width=\textwidth]{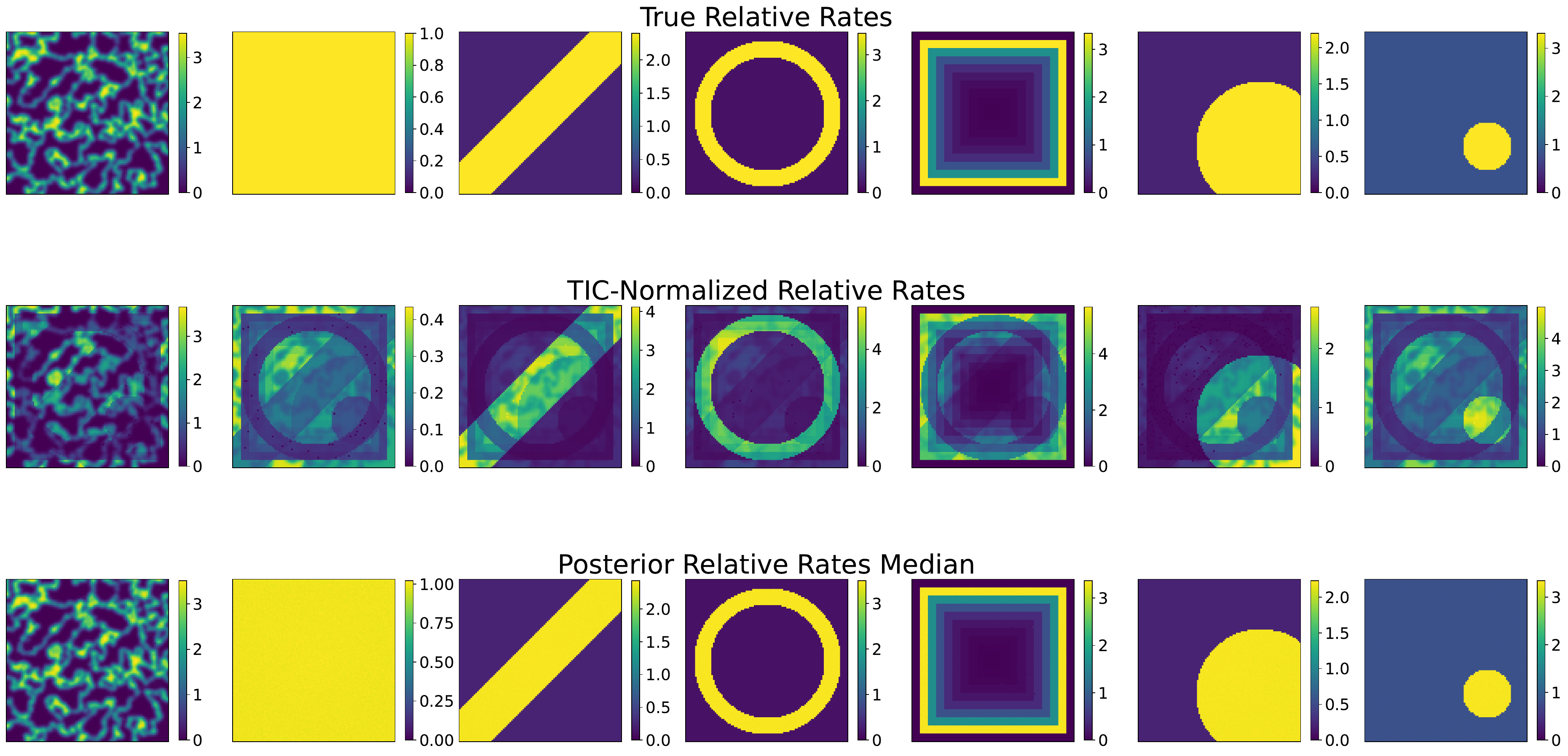}
    \caption{Top row shows true relative rates, middle row shows TIC-normalized relative rates, and bottom row shows posterior medians recovered by our model. Color refers to unitless relative spatial intensities for each metabolite. Posterior medians recover true signal which is convolved in TIC-normalized data.}
    \label{fig:relative_rates}
\end{figure}

We demonstrate the effectiveness of our framework in both simulation and on a real data case study. Simulations clearly show that our approach enables recovery of the true rates, whereas simply reporting the proportion across pixels yields false conclusions on the spatial distribution of a molecule (\cref{fig:relative_rates}). Applying our framework on data from a recent spatial metabolomics study of kidney tissue~\cite{wang:etal:2022:spatial-isotope-tracing} shows provocative differences in the spatial distribution of key metabolites, suggesting our approach is a superior normalization strategy to the current standard in the field.~\looseness=-1


\section{Background and related work}
\label{sec:background}
\subsection{Spatial profiling and compositional data}
\label{subsec:background:spatial-profiling}

Many common methods and assays in spatial biology deal in compositional data measured across a grid of pixels. In spot-based spatial transcriptomics, many deconvolution methods produce relative proportions of cell types in each tissue grid~\cite{miller:etal:2022:reference-free-st-deconv,cable:etal:2022:rctd,lopez:etal:2022:destvi,zhang:etal:2023:bayestme}. Spectral imaging technologies, such as imaging mass spectrometry (IMS) and imaging mass cytometry (IMC), are semiquantitative in nature; while raw molecular counts are technically produced, these only purport to represent relative molecular abundance, rather than absolute abundance. This is compounded by highly variable total abundances measured at each location, restricting spatial analysis in heterogeneous tissues~\cite{Wulff2018-ql}.

For example, in matrix-aided laser desorption ionization time of flight (MALDI-TOF), a common form of IMS in spatial metabolomics,  this is chiefly due to two effects: first, imperfection in both the imaging laser and the co-crystallization of the matrix with the tissue leads to variability in the total amount of molecules that undergo ionization at each location; second, interaction effects, both between metabolite types and between these and the matrix substrate, suppress the ionization of particular molecules. To account for this, biologists will often normalize the observed molecular presence by dividing the total ion count (TIC) within each pixel, a process known as TIC-normalization~ \cite{Alexandrov2020-az,Deininger2011-ga}. 

Furthermore, spatial profiling data are often incomplete, with missing data caused both by random and systemic sources. In particular, some imaging technologies cannot report molecules whose presence is below a certain limit of detection (LOD), leading to left-censoring~\cite{Karpievitch2012-yt,Xie2015-rj}. This is particularly challenging for molecules at low rates, as the LOD threshold may represent substantial censoring of the relative rate. A complete statistical approach to modeling spatial compositional data in biology requires handling these missing-not-at-random data.

\subsection{Related work in graphical models}

Markov Random Field approaches have been used to model graphical networks in spatial problems for decades  \citep{Banerjee2009-lp,Cressie2015-jy}. More recently, sparse modeling techniques such as Nearest Neighbors Gaussian Processes \cite{Datta2016-ae} and Vecchia Approximations \cite{Katzfuss2021-zn} have demonstrated that joint processes can be effectively expressed by sequences of conditional probabilities, allowing for computationally efficient joint probability modeling at large scales.~\looseness=-1

Our work also builds on the use of shrinkage priors in Bayesian applications to capture meaningful signals in complex, high-dimensional environments \cite{Park2008-nm,Wang2012-fc}. In particular, our work is an application of global-local shrinkage priors, such as the Horseshoe \cite{Carvalho2009-ap} and the gamma-lasso \cite{Taddy2010-lm} (or equivalently double Pareto \cite{Armagan2013-li}) priors, the last of which adds long tails to the traditional lasso shrinkage allowing for a data-driven approach to identifying how much meaningful signals should deviate from zero.~\looseness=-1

\subsection{Related work in hierarchical Variational Inference}

For inference, our method relies on a well-established body of work on variational inference (VI) over the past decade. VI is a widely used technique for approximating posterior distributions in Bayesian inference through a set of "variational" distributions (collectively known as the variational family) \citep{Jordan1999-yl}. This is done by maximizing the ELBO, a lower bound on the KL-divergence between the posterior distribution and the variational family. Expanding on this, Stochastic Variational Inference was introduced as a method to obtain approximate Bayesian posteriors for large datasets \citep{Hoffman2012-cl}. Further advances in VI were made with Automatic Differentiation Variational Inference (ADVI) \citep{Kucukelbir2016-fp}, which leverages existing autodifferentiation technology to calculate stochastic gradients with minimal effort, and reparametrized sampling \citep{PKingma2013-la}, which can efficiently calculate intractable expectations through differentiable MCMC methods. Recent advances in less trivial reparameterized sampling for continuous distributions have further allowed for more expressive variational families of the kinds we employ in this work \citep{Mohamed2019-ar}. ~\looseness=-1

Variational Inference techniques generally leverage independence wherever possible, for tractability and computational efficiency, and the most common approach remains the mean-field variational family where variational distributions are assumed independent across model components \citep{Blei2017-py}. However, mean-field variational techniques are known to be limited when approximating posteriors over highly dependent joint distributions, including in spatial settings \citep{Wu2022-mn,Ren2011-wj}. To that end, significant work has been done in developing more expressive approximate distributions which impart structure onto the variational family. For example, hierarchical variational methods \citep{Ranganath2015-gg} introduce shared priors to couple variational distributions, while structured variational inference \citep{pmlr-v38-hoffman15} imposes conditional dependencies among local variables. It has been shown that such structured variational families often out perform full-rank approaches while sacrificing minimal computational burden over mean-field VI \citep{JoohwanKo}. ~\looseness=-1

In spatial problems, a common approach has been approximating posteriors with a low-rank multivariate normal distribution, with compelling and competitive results \citep{Hensman-bigdata13}. Nevertheless, these approaches have been known to struggle in capturing low-length scale changes of the kind we expect with piecewise constant patterns in biological imaging data. We instead adopt a sparse structured variational approach, placing conditionally independent variational distributions over nodes dependent on distributions over edges. Thus, we maintain conceptual coherence in using a sparse variational distribution over a sparse posterior. In this, we echo  recent work in the development of a sparse variational approach for NNGPs \citep{Wu2022-mn}.~\looseness=-1

\section{Model: The censored graph-fused gamma lasso}
\label{sec:model}

\subsection{Notation}
\label{subsec:model:notation}

We consider molecular mass observed across a grid of pixels in a tissue, which we formalize as an undirected graph $\mathcal{G}=\{\mathcal{E},\mathcal{V}\}$ of edges $\mathcal{E}$ and vertices $\mathcal{V} $, with the number of edges denoted $R=|\mathcal{E}|$ and the number of vertices denoted denoted $M=|\mathcal{V}|$. We will denote $e_{i,j}$ as the edge which connects vertices $i$ and $j$. Let $x_{i,d}$ be molecules observed at vertex (also referred to as location) $i \in \{1,...,M\}$ of type $d \in \{1,...,D\}$. Let $\textbf{x}_i$ refer to the $D$-sized vector of molecular counts at location $i$. 
Further, define total $N_i = \sum_{j=1}^D x_{ij}$. Let $p_{i,d} = \frac{\theta_{i,d}}{\sum_{k=1}^D \theta_{i,k}}$, where $\theta_{i,d}$ is the latent molecular rate. Let $\bs{\theta}_d$ refer to the size $M$ vector of latent rates for molecular type $d$ across all locations. Lastly, our target variable of interest is $\tilde{\bs{\theta}}_d=\frac{\bs{\theta}_d}{||\bs{\theta}_d||_{\ell_1}}$, the molecular rate normalized across all locations. 

\subsection{Multinomial likelihood with censored total}
\label{subsec:model:likelihood}

To model competitive sampling in spatial profiling data, we use a multinomial model over the observed counts. The total $N_i$ represents the number of molecules detected by the biological assay at location $i$ with molecular rate vector $\mathbf{p}_i$,
\begin{equation}
\begin{aligned}
\label{eqn:likelihood}
[\textbf{x}_1, ...,\textbf{x}_m|\bs{\theta}_1,...,\bs{\theta}_d] &\sim \prod_{i=1}^M Mult(\textbf{p}_i; N_i) \, , &
p_{i,d} &= \frac{\theta_{i,d}}{\sum_{d=1}^D \theta_{i,d}}
\end{aligned}
\end{equation}


A natural extension to  address the left-censored data would be to draw on Bayesian Survival Analysis techniques~\citep{Ibrahim2001-my}, which model censored data with the CDF of the likelihood. However, left-censoring in this context creates a problem  for a straightforward implementation of a multinomial likelihood model: at any location with censoring on any specific molecule, the overall total molecule count is only partially observed, implying the multinomial cannot be fully parameterized.

To overcome this, we turn to the negative multinomial distribution as an extension of the multinomial for an unknown total. The negative multinomial, a multivariate augmentation of the negative binomial, models the number of successes across a range of outcomes given a total number of failures, specified by a set of probabilities whose total sum remains less than 1. In our context, we model the observed counts marginally as a multinomial and the censored counts as a negative multinomial, conditional on the total observed counts; both of these distributions are parameterized by the same set of underlying probabilities. Thus our joint likelihood over the data — observed and censored — is as follows:
\begin{equation}
\label{eqn:likelihood-negative-multinomial}
\begin{split}
\prod_{i=1}^M P(\textbf{x}_i^O, \textbf{x}_i^C|\textbf{p}_i) & =  \prod_{i=1}^MP(\textbf{x}_i^C|\textbf{x}_i^O, \textbf{p}_i) P(\textbf{x}_i^O| \textbf{p}_i)\\
&= \prod_{i=1}^M \Psi (LOD(d \in \mathcal{C}_i); \textbf{p}_i^C, N_i) \times Mult(\textbf{x}_i^O; \frac{{\textbf{p}}^O_i}{||{\textbf{p}}^O_i||_{\ell_1}}, N_i)
\end{split}
\end{equation}

Here, the superscripts $O$ and $C$ refer to observed and censored molecules, and $\Psi$ refers to the CDF of a negative multinomial distribution evaluated at each of the limits-of-detection for set of the censored molecular types $\mathcal{C}_i$. No analytic form exists for this CDF, so $\Psi$ is calculated via an efficient Monte Carlo sampling scheme~\citep{Zhou2022} (see Appendix for details).



\subsection{Graph-fused gamma lasso prior}
\label{subsec:model:prior}

To enforce data-dependent sparsity between rates of the same molecule in adjacent locations, we use a heavy-tailed variant of the graph-fused lasso prior~\citep{Tansey2017-ub} on the latent log-rates $\log \bs{\theta}$ across the spatial graph. Specifically, we penalize absolute differences in locally adjacent rates subject to an edge- and molecule type-specific shrinkage parameter. Defining $\xi(i): \{j \in \mathcal{V} | e_{i,j} \in \mathcal{E}\}$, we
we impose independent gamma-Laplace priors along the edges of graph $\mathcal{G}$. For each edge $e_{i,j}\in\mathcal{E}$, we define the edge-adjacency matrix $\textbf{H}_{R\times M}$, where each row denotes an edge and each column denotes a vertex,
$$
\textbf{H}_{e,t} = \begin{cases}
1, \ t=i\\
-1, \ t=j\\
0, \ \text{otherwise}
\end{cases}
$$
We then apply the sparsity-inducing prior to the transformation of the log-rates,
\begin{equation}
\begin{aligned}
\label{eqn:gfgl-prior}
    \boldsymbol{\alpha}_d=\textbf{H}\log\boldsymbol{\theta}_d & \quad & \alpha_{r,d} \sim Laplace(0,1/\nu_{r,d}) & \quad &  \nu_{r,d}\sim Exp(\lambda_d),
\end{aligned}
\end{equation}
where $r\in\{1,...,R\}$ is the edge index and $d\in\{1,...,D\}$ is the molecule index.
We use a scale-mixture-of-normals decomposition to specify the compound gamma-Laplace and to aid inference~\cite{polson:scott:2010:shrink-globally}. This separates the prior into global ($\lambda_d$) and local ($\nu_{r,d}$) shrinkage  terms, enabling a flexible graphical shrinkage approach that forces adjacent differences to zero but allows for large deviations at local change-points. For a fully Bayesian approach, $\lambda_d$ itself is given an Exponential prior with global shrinkage hyperparameter $\tau_d$.~\looseness=-1


Given the transformation in \cref{eqn:likelihood}, which corresponds to the softmax function on $\log \bs{\theta}$, this model is only identifiable in $\bs{\theta}$ up to some multiplicative constant (or equivalently an additive constant in $\log \bs{\theta}$). However $\tilde{\bs{\theta}}$ is also invariant to a multiplicative shift, so we retain identifiability in our parameter of interest. In particular, we can think of the graph fusion prior as effectively shrinking the parameter space of an otherwise underdetermined system. A set of $M$ independent $D$-sized multinomial distributions has $M\times (D-1)$ degrees of freedom, and therefore  is unidentifiable at $M\times D$ parameters. However it is known that the degrees of freedom of a fused lasso is equivalent to the number of change points across the graph \citep{10.1214/12-AOS1003}. Therefore, so long as the number of change points is sufficiently small (specifically no more than $M-M/D$), we reduce the parameter space to $M\times (D-1)$ or fewer dimensions, restoring identifiability. In this sense, the prior acts analogously to sparsity-inducing priors in the compressed sensing literature, which allow for recovery of high-dimensional signals from underdetermined observations by exploiting low-dimensional structure~\cite{8260873}.

\section{Inference: ADVI with sparse structured variational distribution}
\label{sec:inference}

Our key variable of interest is the joint posterior  $P( 
\tilde{\theta}_{1,1},...,\tilde{\theta}_{D,M}|\cdot)$ over the spatially-normalized latent metabolic rates. While the overall Bayesian structure naturally lends itself to  MCMC inference, we turn to variational inference for a computationally efficient approach to estimating approximate posteriors.
\begin{figure}[t!]
    \centering
    \resizebox{0.95\textwidth}{!}{   \usetikzlibrary{arrows.meta, fit, positioning}

\begin{tikzpicture}[
    node distance=0.7cm and 1.8cm,
    circ/.style={circle, draw, minimum size=1.2cm, align=center, line width=1pt},
    ->, >=Stealth, line width=1.5pt,
    font=\large\bfseries,
    panel/.style={draw, thick, rounded corners, inner sep=0.5cm, line width=1pt},
    labelbox/.style={fill=white, inner sep=2pt, font=\large\bfseries},
    rightpanel/.style={draw, thick, rounded corners, minimum width=5cm, minimum height=7cm, line width=1pt}  
]

\scope[xshift=-8cm]
\node (a1) {\(a^*_{e_{ij}}\)};
\node[below=of a1] (b1) {\(b^*_{e_{ij}}\)};
\node[circ, right=of a1, yshift=-0.7cm] (q1) {\(q(\nu_{e_{ij}})\)};
\draw[->] (a1.east) -- (q1.west);
\draw[->] (b1.east) -- (q1.west);

\node[below=1.8cm of b1] (a2) {\(a^*_{e_{ji'}}\)};
\node[below=of a2] (b2) {\(b^*_{e_{ji'}}\)};
\node[circ, right=of a2, yshift=-0.7cm] (q2) {\(q(\nu_{e_{ji'}})\)};
\draw[->] (a2.east) -- (q2.west);
\draw[->] (b2.east) -- (q2.west);

\node[below=1.8cm of b2] (a3) {\(a^*_{e_{i'j'}}\)};
\node[below=of a3] (b3) {\(b^*_{e_{i'j'}}\)};
\node[circ, right=of a3, yshift=-0.7cm] (q3) {\(q(\nu_{e_{i'j'}})\)};
\draw[->] (a3.east) -- (q3.west);
\draw[->] (b3.east) -- (q3.west);

\node[right=4cm of q2] (sigma2j) {\(\sigma^{2*}_j\)};
\node[above=0.5cm of sigma2j] (mu_j) {\(\mu_j^*\)};
\node[circ, right=of sigma2j] (theta_j) {\(q(\theta_j)\)};
\draw[->] (mu_j) -- (theta_j);
\draw[->] (sigma2j) -- (theta_j);

\node[below=3.9cm of sigma2j] (sigma2i) {\(\sigma^{2*}_{i'}\)};
\node[above=0.5cm of sigma2i] (mu_i) {\(\mu_{i'}^*\)};
\node[circ, right=of sigma2i] (theta_i) {\(q(\theta_{i'})\)};
\draw[->] (mu_i) -- (theta_i);
\draw[->] (sigma2i) -- (theta_i);

\node[fit=(a1)(b3)(q1)(q2)(q3), panel, label=above:{\textbf{Variational Distribution over $\boldsymbol{\nu}$}}] (panel1a) {};
\node[fit=(mu_j)(theta_j)(mu_i)(theta_i), rightpanel, label=above:{\textbf{Variational Distribution over $\boldsymbol{\theta}$}}] (panel2a) {};

\node[above=1.0cm of panel1a] {\LARGE \textbf{Mean Field Variational Inference}};
\endscope

\scope[xshift=8cm]
\node (a1) {\(a^*_{e_{ij}}\)};
\node[below=of a1] (b1) {\(b^*_{e_{ij}}\)};
\node[circ, right=of a1, yshift=-0.7cm] (q1) {\(q(\nu_{e_{ij}})\)};
\draw[->] (a1.east) -- (q1.west);
\draw[->] (b1.east) -- (q1.west);

\node[below=1.8cm of b1] (a2) {\(a^*_{e_{ji'}}\)};
\node[below=of a2] (b2) {\(b^*_{e_{ji'}}\)};
\node[circ, right=of a2, yshift=-0.7cm] (q2) {\(q(\nu_{e_{ji'}})\)};
\draw[->] (a2.east) -- (q2.west);
\draw[->] (b2.east) -- (q2.west);

\node[below=1.8cm of b2] (a3) {\(a^*_{e_{i'j'}}\)};
\node[below=of a3] (b3) {\(b^*_{e_{i'j'}}\)};
\node[circ, right=of a3, yshift=-0.9cm] (q3) {\(q(\nu_{e_{i'j'}})\)};
\draw[->] (a3.east) -- (q3.west);
\draw[->] (b3.east) -- (q3.west);

\node[right=4cm of q2] (gammaj) {\(\gamma_j\)};
\node[above=0.5cm of gammaj] (muj) {\(\mu_j^*\)};
\node[circ, right=of gammaj] (theta_j) {\(q(\theta_j)\)};
\draw[->] (q1.east) to[out=10,in=160] (gammaj.north west);
\draw[->] (q2.east) to[out=0,in=180] (gammaj.west);
\draw[->] (muj.east) -- (theta_j.north west);
\draw[->] (gammaj.east) -- (theta_j.west);

\node[below=3.7cm of gammaj] (gammai) {\(\gamma_{i'}\)};
\node[above=0.5cm of gammai] (mui) {\(\mu_{i'}^*\)};
\node[circ, right=of gammai] (theta_i) {\(q(\theta_{i'})\)};
\draw[->] (q2.east) to[out=-10,in=160] (gammai.north west);
\draw[->] (q3.east) to[out=0,in=180] (gammai.west);
\draw[->] (mui.east) -- (theta_i.north west);
\draw[->] (gammai.east) -- (theta_i.west);

\node[fit=(a1)(b3)(q1)(q2)(q3), panel, label=above:{\textbf{Variational Distribution over $\boldsymbol{\nu}$}}] (panel1b) {};
\node[fit=(muj)(theta_j)(mui)(theta_i), rightpanel, label=above:{\textbf{Variational Distribution over $\boldsymbol{\theta}$}}] (panel2b) {};

\node[above=1.0cm of panel1b] {\LARGE \textbf{Sparse Structured Variational Inference}};
\endscope

\end{tikzpicture}}
    \caption{Comparison of mean field and sparse structured Variational Inference. Left: mean-field assumes independence between latent variables. Right: structured inference models dependencies via shared latent contributions.}
    \label{fig:vi_diagrams}
\end{figure}
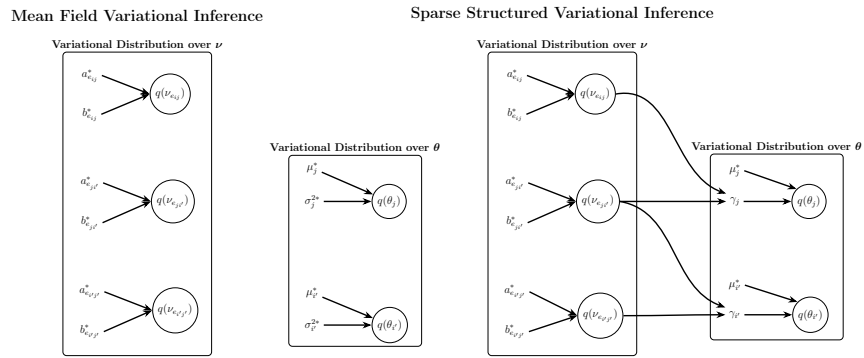

\subsection{Sparse structured variational family}
\label{subsec:inference:VI_family}

While the model priors are defined across the graph edges, the variable of interest remains defined over the vertices. As a result, our variational family entails a distribution on $\log\bs{\theta}_d$ rather than $\bs{\alpha}_d$ (along with distributions over $\bs{\nu}_d$ and $\lambda_d$). Given the well known problems with mean-field VI in capturing joint posteriors, we developed a hierarchical, conditionally independent variational family which leverages sparsity in the same manner as the graph-fused gamma lasso prior. \Cref{fig:vi_diagrams} details a schematic comparing our VI approach with mean-field over a set of 3 edges and 2 nodes. In particular, by parameterizing a distribution across $\log\bs{\theta}_d$ dependent on a vertex-transformed distribution over $\bs{\nu}_d$, we create a set of spatially-dependent marginal variational distributions on $\log\bs{\theta}_d$ while maintaining conditional independence for sampling and gradient estimation. To that end, we introduce: \begin{equation}
    \bs{\gamma}_d =(\textbf{H}^+)^T(\bs{\nu}_d^{-1})
\end{equation} where $h^+_{r,i} = |h_{r,i}|\ \ \forall r\in\{1,..,R\} \text{ and } i\in\{1,..,M\}$, and $\bs{\nu}_d^{-1}=(1/\nu_{1,d},...,1/\nu_{R,d})^T$. Therefore $\bs{\gamma}_d$ is an $M$ length vector where each entry is the sum of the modeled variability across connecting edges.
We maintain mean-field variational distributions over $\lambda_1,..,\lambda_D$ and $\bs{\nu}_1,...,\bs{\nu}_D$, completing our variational family over factorized distributions:

\begin{align}
Q(\lambda_1,...,\lambda_D) &= \prod_{d=1}^D \Gamma(\lambda_{0d}^*,\lambda_{1d}^*)\\
Q(\bs{\nu}_1,...,\bs{\nu}_D) &= \prod_{d=1}^D \prod_{r=1}^R \Gamma(a_{r,d}^*, b_{r,d}^*)\\
Q(\log \bs{\theta}_1,...,\log \bs{\theta}_D) &= \prod_{d=1}^D \prod_{i=1}^M \int N(\mu^*_{i,d}, \gamma_{i,d}^{-1}) Q( \gamma_{i,d})d \gamma_{i,d} \nonumber\\
&= \prod_{d=1}^D \prod_{i=1}^M 
\int N(\mu^*_{i,d}, (\sum_{r \in \xi(i)} \nu_{r,d}^{-1})^{-1}) 
\prod_{r \in \xi(i)} Q(\nu_{r,d}) \, d\nu_{r,d}
\end{align}

We optimize the ELBO with respect to $\mu_{i,d}^*$,  $a_{r,d}^*$, $b_{r,d}^*$, $\lambda_{0d}^*$, and $\lambda_{1d}^*$. This contrasts with a fully factorized mean-field approach, which would simply be
\begin{equation}
Q(\log \bs{\theta}_1,...,\log \bs{\theta}_D) = \prod_{d=1}^D \prod_{i=1}^M N(\mu^*_{i,d},(\sigma^*_{i,d})^2) \, ,
\end{equation}
adding $\sigma^*_{i,d}$ to the set of variational parameters.

\subsection{Gradient calculation}  
\label{subsec:inference:gradient}

The specific ELBO function we are maximizing is
\begin{equation}
\begin{split}
    &\sum_{i=1}^M\{ E_{Q(\bs{\Theta})}[\log \mathcal{L}(\textbf{x}_i|\textbf{p}_i)] \}
    +\sum_{d=1}^D\{E_{Q(\bs{\theta}_d,\bs{\nu}_d)}[\log P(\bs{\alpha_d})] + E_{Q(\bs{\nu}_d)}[\log P(\bs{\nu_d})] \\
    &- E_{Q(\bs{\theta}_d,\bs{\nu}_d)}[\log Q(\bs{\theta}_d|\bs{\nu}_d)] - E_{Q(\bs{\nu}_d)}[\log Q(\bs{\nu_d})] \\
    &+E_{Q(\lambda_d)}[\log P (\lambda_d)] -E_{Q(\lambda_d)}[\log Q (\lambda_d)]\} \, .
\end{split}
\end{equation}
Here, $\mathcal{L}$ represents the likelihood in \cref{eqn:likelihood-negative-multinomial} taken over the set of all rates $\bs{\Theta}$. Recall that $\textbf{p}$ and $\bs{\alpha}$ are connected to $\bs{\theta}$ by the transformations in \cref{eqn:likelihood} and \cref{eqn:gfgl-prior}, respectively. Implicit gradients with respect to the individual ELBO expectations are calculated with automatic differentiation through reparameterized sampling and subsequently evaluating the function of interest over the samples. Due to the sparse conditional prior in \cref{eqn:gfgl-prior}, reparameterized sampling gradients would still be required with a fully factorized mean-field approach. As such, our variational family imparts minimal computational burden on inference with respect to mean-field VI. \Cref{alg:ELBO} provides an outline of how the ELBO is calculated. Full algorithmic details on the inference can be found in the Appendix.
\begin{algorithm}
\caption{Reparametarized Sampling ELBO Calculation for HV-GFGL with Left-Censoring}
Where $O(i,d) = 1$ if $x_{i,d}$ is observed, $0$ otherwise, and $s\in \{1,...,S\}$ is the number of samples:
\label{alg:ELBO}
\begin{enumerate}
    \item 
    $
    \lambda_d^{(s)} \sim Q(\lambda_d), \quad 
    \nu_{r,d}^{(s)} \sim Q(\nu_{r,d}), \quad \gamma_{i,d}^{(s)} = \sum_{j \in \xi(i)} \frac{1}{\nu_{e_{i,j},d}^{(s)}}, \quad 
    \log \theta_{i,d}^{(s)} \sim Q(\theta_{i,d})
    $
    
    \item 
    $p_{i,d}^{(s)} = \frac{\theta_{i,d}^{(s)}}{\sum_d \theta_{i,d}^{(s)}}, \quad 
    \tilde{p}_{i,d}^{(s)} = \frac{p_{i,d}^{(s)} O(i,d)}{\sum_d p_{i,d}^{(s)} O(i,d)}
   $
    
    \item 
   $
    ELBO = \frac{1}{S} \sum_{i,d,s} x_{i,d} \log \tilde{p}_{i,d}^{(s)} + 
    \frac{1}{S} \sum_{i,s} \log \Psi\left[\text{LOD}(d \in \mathcal{C}_i); 
    (\mathbf{p}^C_i)^{(s)}, N_i\right]
   $
    
    \item 
    $
    ELBO += \frac{1}{S^2} \sum_{i,d,s_1,s_2} \sum_{j \in \xi(i)} 
    \frac{-|\log \theta_{i,d}^{(s_2)} - \log \theta_{j,d}^{(s_2)}|}
    {\nu_{e_{i,j},d}^{(s_1)}} -
    \frac{1}{S} \sum_{i,d,s}(-\log \gamma_{i,d}^{(s)})
    $
    
    \item 
   $
    ELBO += -\frac{1}{S} \sum_{d,s} \text{KL}(Q(\bs{\nu}_d)|| 
    \text{Exp}(\lambda_d^{(s)})) - 
    \sum_d \text{KL}(Q(\lambda_d)|| \text{Exp}(\tau_d))
   $
\end{enumerate}

\end{algorithm}

\subsection{Hyperparameters and initialization}
\label{subsec:inference:initialization}

Due to the nonconvexity of the ELBO function, variational inference is known to be sensitive to choices of initialization and hyperparameters \cite{Blei2017-py}. In this model, the critical choices are in the initialization of $b_{r,d}^*$, the rate parameter of $\nu_{r,d}$ which controls both the local shrinkage and the posterior variance of $\log\theta_{i,d}$, along with the initialization of $\lambda_{1d}^*$ and the choice of the hyperparameter $\tau_d$, which control the global shrinkage.

We adopt an empirical Bayesian approach following a few heuristics. First, parameters should be initialized to the scale of the molecular data, otherwise convergence to a local optimum with over- or under- shrinkage is nearly guaranteed. Second, the posterior variance of the rates should increase proportionally to its mean, as is typical with count data. Last, molecules with greater abundance should require stronger shrinkage parameters for equally sparse spatial signals. To this end, we initialize $b_{r,d}^*=E(\textbf{x}_d)$ for all edges $r$. Since $E(\nu_{r,d}) \approx \lambda_{1d}^*$, we also initialize $ \lambda_{1d}^*=E(\textbf{x}_d)$. We initialize $\tau_d=Var(\textbf{x}_d)$, reflecting our uncertainty in the global shrinkage level. This allows the model to adapt more flexibly to heterogeneous spatial patterns, reflecting the belief that more variable molecules may contain more complex spatial structure and should be afforded greater flexibility.

We initialize $\mu^*_{i,d}$ to the local proportion (which is equivalent to the maximum likelihood estimate assuming independent multinomials). $\lambda^*_{0d}$ and $a^*_{r,d}$ are initialized at 1 and 2, respectively, for all indices, allowing sparsity in the local shrinkage but preventing the posterior variance of the rates from collapsing to zero. In both simulation and real-world applications, all of these choices were shown to accelerate convergence, ensure numerical stability, and increase posterior accuracy.

\section{Simulation Study}

\subsection{Simulation design}

A simulation of MALDI-ToF IMS was conducted to compare our methodology against the state-of-the-practice in IMS (TIC-normalization), as well as a set of simpler models. Data were simulated to emulate  biological structures with varying degrees of heterogeneity and left-censoring across seven "metabolites" (denoted metabolite one, metabolite two, etc.). Figure \ref{fig:relative_rates} illustrates the relative true states and relative TIC values, along with the eventual posterior median of the relative rates recovered by our model. The Appendix details specifics on the simulation implementation.~\looseness=-1

While current methodologies for recovering relative rates under competitive sampling in spatial biological data are limited, we compared our hierarchical variational graph-fused gamma lasso approach (denoted HV-GFGL)  with more readily applicable mean-field VI methods. We implemented two additional models which reduce some of the complexity of our suggested approach. Firstly, we implemented the same GFGL prior but with a fully factorized mean-field variational family. We also implement a mean-field VI with a standard graph fused lasso prior. These models are denoted MF-GFGL and MF-GFL, respectively. ~\looseness=-1

All three models were implemented in Pytorch using Nvidia T4 GPUs in a Google Colab environment. Each model was run to $25,000$ iterations, a number chosen to ensure convergence. Both GFGL models ran at  $\approx0.01$ seconds per iteration, while the GFL model ran at $\approx0.008$ seconds per iteration. The number of samples for the negative multinomial CDF calculation was set to $100$, while the number of samples for the re-parameterized gradient was set to $2$. We set hyperparameters and initializations according to \cref{subsec:inference:initialization}. ~\looseness=-1

\subsection{Results}
 RMSE with respect to relative rates for our suggested model, the two benchmark models, and TIC-normalization is shown in~\cref{tab:rmse_results}.  The GFGL  models  improve upon TIC-normalization in RMSE by 1 to 2 orders of magnitude across all metabolites. The simple fused lasso performs similiarly except in metabolite five, where it has an RMSE more than twice TIC. Since metabolite five has a more heterogeneous pattern, this suggests that the more flexible global-local shrinkage approach is appropriate for more complicated patterns. 

The real strength of our sparse structured VI approach is exhibited in~\cref{tab:ci_coverage_split}, which shows $90 \%$ and $50\%$ credible interval coverage for the three models. The two benchmark models exhibit the common problem of severe posterior under-coverage. On the other hand, our methodology has overall credible interval coverage  of $0.86$ and $0.51$ for  $90\%$ and $50\%$ CIs respectively, reflecting a significantly better calibrated and approximated posterior.

\begin{table}[ht]
\centering
\caption{Root Mean Squared Error (RMSE) for TIC and three models across seven metabolites and overall. Best values per column bolded.}
\begin{tabular}{lcccccccc}
\toprule
Model & RMSE$_1$ & RMSE$_2$ & RMSE$_3$ & RMSE$_4$ & RMSE$_5$ & RMSE$_6$ & RMSE$_7$ & Overall \\
\midrule
TIC     & 0.6256 & 0.5946 & 0.4865 & 0.4807 & 0.4856 & 0.6443 & 0.6527 & 0.5719\\
HV-GFGL & \textbf{0.0239} & 0.0055 & 0.0051 & 0.0056 & \textbf{0.0065} & 0.0204 & 0.0038 & 0.0127 \\
MF-GFGL & 0.0296 & \textbf{0.0009} & \textbf{0.0012} & \textbf{0.0015} & 0.0116 & \textbf{0.0012} & \textbf{0.0010} & \textbf{0.0120} \\
MF-GFL  & 0.0727 & 0.0045 & 0.0116 & 0.0227 & 1.0741 & 0.0167 & 0.0346 & 0.4073 \\
\bottomrule
\end{tabular}
\label{tab:rmse_results}
\end{table}
\begin{table}[ht]
\centering
\caption{90\% Credible Interval (CI) coverage for three models across seven metabolites and overall. 50\% CI coverage shown in parentheses. Best 90\% values in each column are bolded.}
\begin{tabular}{lcccc}
\toprule
Model & CI$_1$ & CI$_2$ & CI$_3$ & CI$_4$ \\
\midrule
HV-GFGL&\textbf{0.74} (0.52) & \textbf{0.99} (0.71) & \textbf{0.88} (0.50) &\textbf{0.86} (0.47)  \\
MF-GFGL  & 0.14 (0.06) & 0.27 (0.11) & 0.13 (0.05) & 0.04 (0.10) \\
MF-GFL  & 0.13 (0.05) & 0.02 (0.01) & 0.03 (0.01) & 0.01 (0.00) \\
\bottomrule
\end{tabular}

\vspace{1em}

\begin{tabular}{lcccc}

\toprule
Model & CI$_5$ & CI$_6$ & CI$_7$ & Overall \\
\midrule
HV-GFGL&\textbf{0.69} (0.39)  & \textbf{0.92} (0.53) & \textbf{0.88} (0.48)& \textbf{0.85} (0.51)\\
MF-GFGL  & 0.00 (0.00) & 0.14 (0.06) & 0.11 (0.05) & 0.13 (0.05) \\
MF-GFL  & 0.00 (0.00)  & 0.01 (0.01) & 0.00 (0.01) & 0.03 (0.01) \\
\bottomrule
\end{tabular}

\label{tab:ci_coverage_split}
\end{table}

\section{Case study}
\subsection{Data description and implementation}
We compare our HV-GFGL model against TIC-normalization in a case study of spatial isotope tracing data on mouse kidneys~\cite{wang:etal:2022:spatial-isotope-tracing}. In these data, nutrient abundance patterns are known to vary substantially across anatomically distinct regions of the kidney, with higher blood flow in the outer cortex contrasting with the inner, more hypoxic medulla where urine is concentrated~\cite{wang:etal:2010:specific-metabolic-rates}. Further, a ring of blood vessels known as the outer stripe separates the cortex from the medulla and is also known to be metabolically distinct.

Ion counts on 349 unique metabolites across 15403 pixels were collected from a mouse kidney using MALDI-TOF IMS. As is typical with IMS data, the intensity of metabolic presence varies widely by metabolite. The raw counts range from from $3.11 \times 10^4$ to $3.35\times 10^8$ and the 10 metabolites with the highest abundance account for $54\%$ of total counts. Similarly, 138 metabolites had less than $15\%$ left-censoring while 67 exhibited greater than $85\%$ left-censoring. 
Due to the high percentage of censoring on certain metabolites, the number of negative multinomial CDF samples was set to 10 for memory management. Hyperparameters and initializations were set according to \cref{subsec:inference:initialization}
, while the number of samples for gradient calculation was set to 2.
\begin{figure}[htbp]
    \centering
    \begin{subfigure}[b]{0.48\textwidth}
        \centering
        \includegraphics[width=\textwidth]{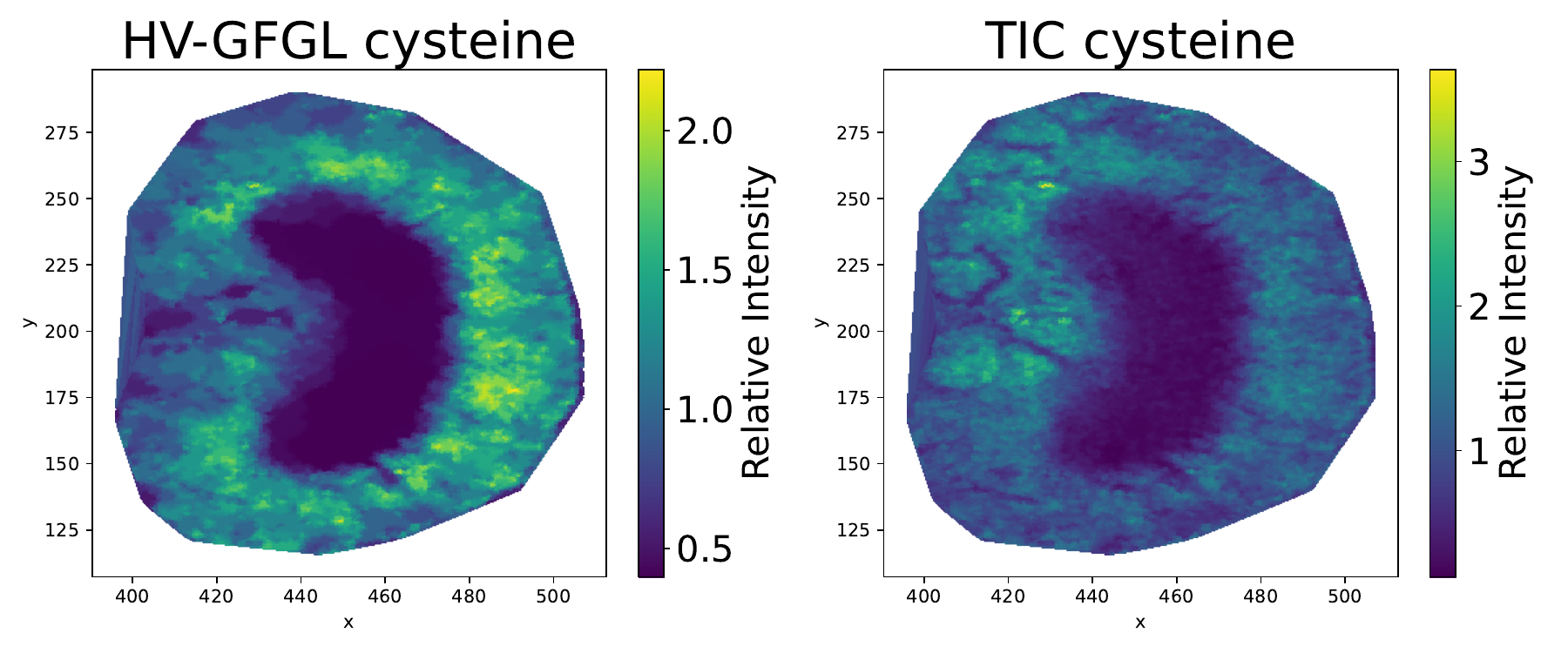}
        \caption{}
        \label{fig:cysteine}
    \end{subfigure}
    \hfill
    \begin{subfigure}[b]{0.48\textwidth}
        \centering
        \includegraphics[width=\textwidth]{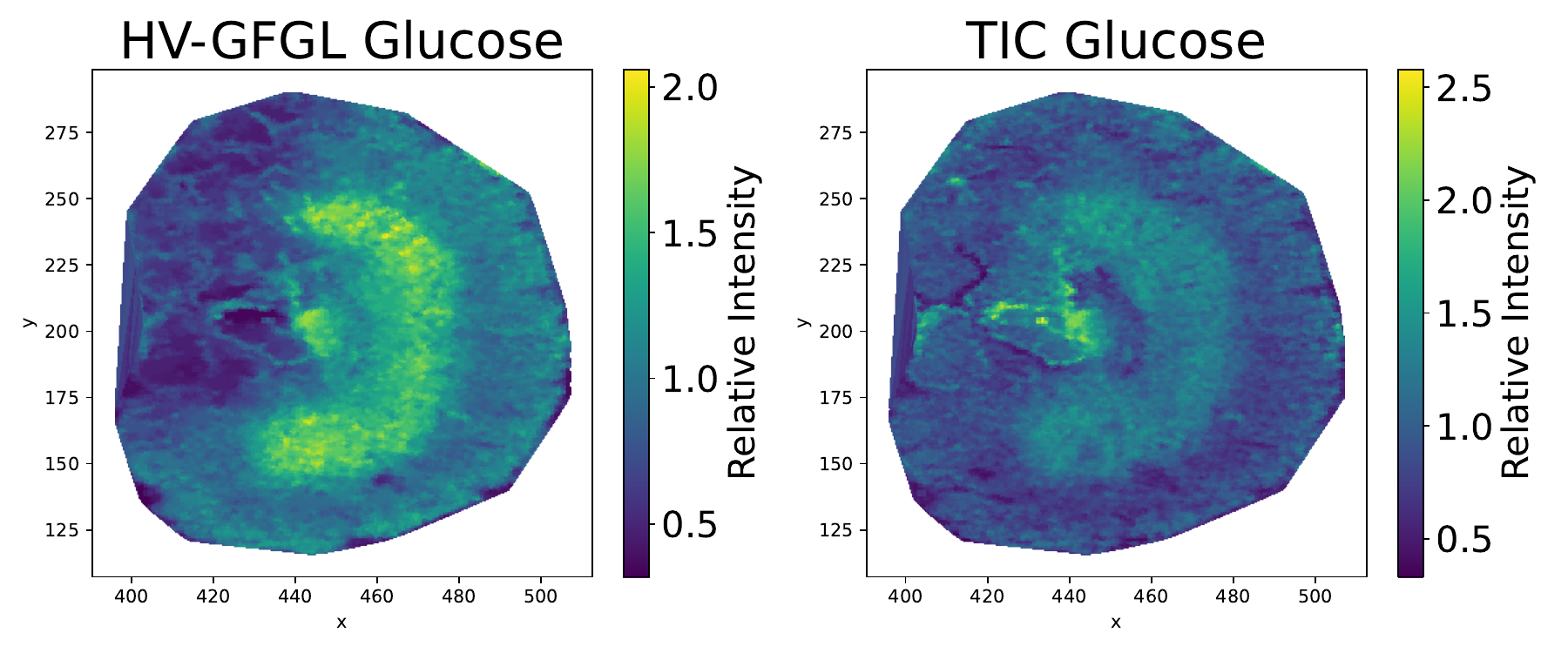}
        \caption{}
        \label{fig:glucose}
    \end{subfigure}

    \vspace{1em}

    \begin{subfigure}[b]{0.48\textwidth}
        \centering
        \includegraphics[width=\textwidth]{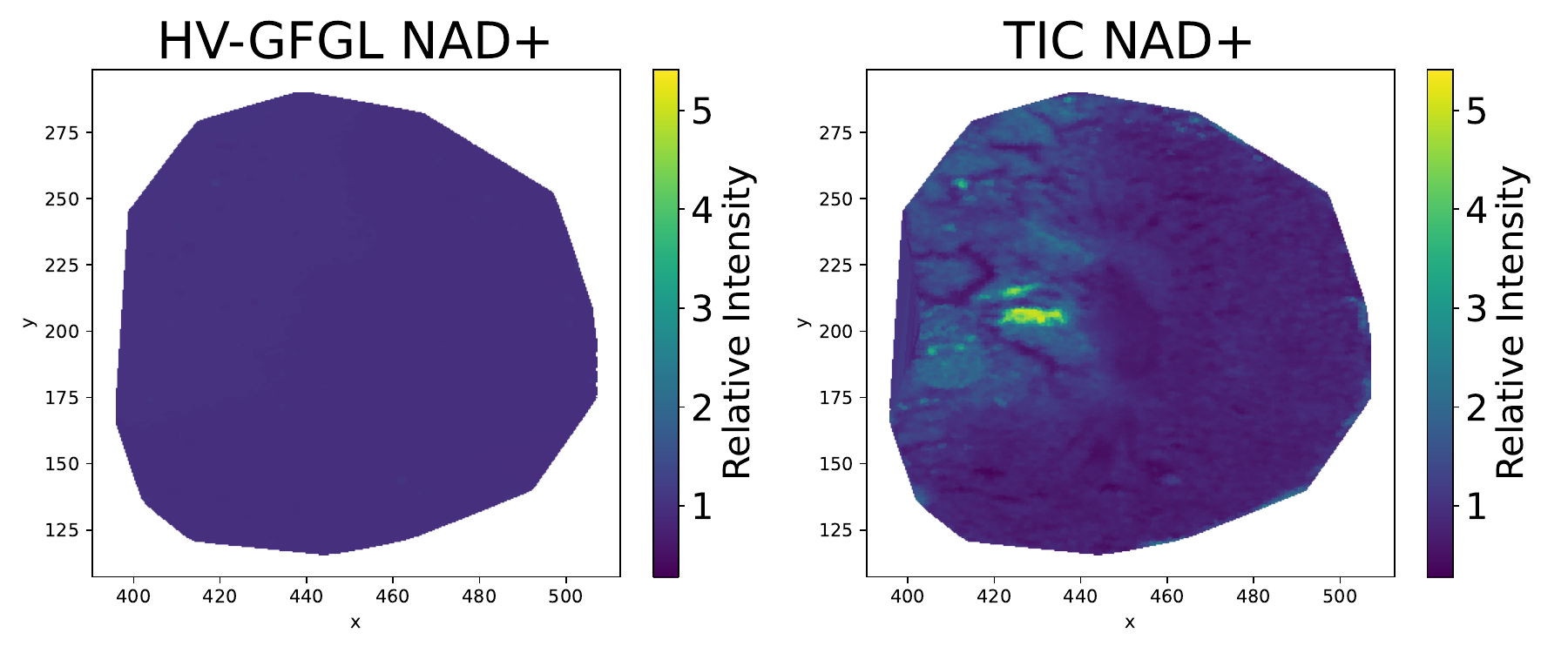}
        \caption{}
        \label{fig:NAD}
    \end{subfigure}
    \hfill
    \begin{subfigure}[b]{0.48\textwidth}
        \centering
        \includegraphics[width=\textwidth]{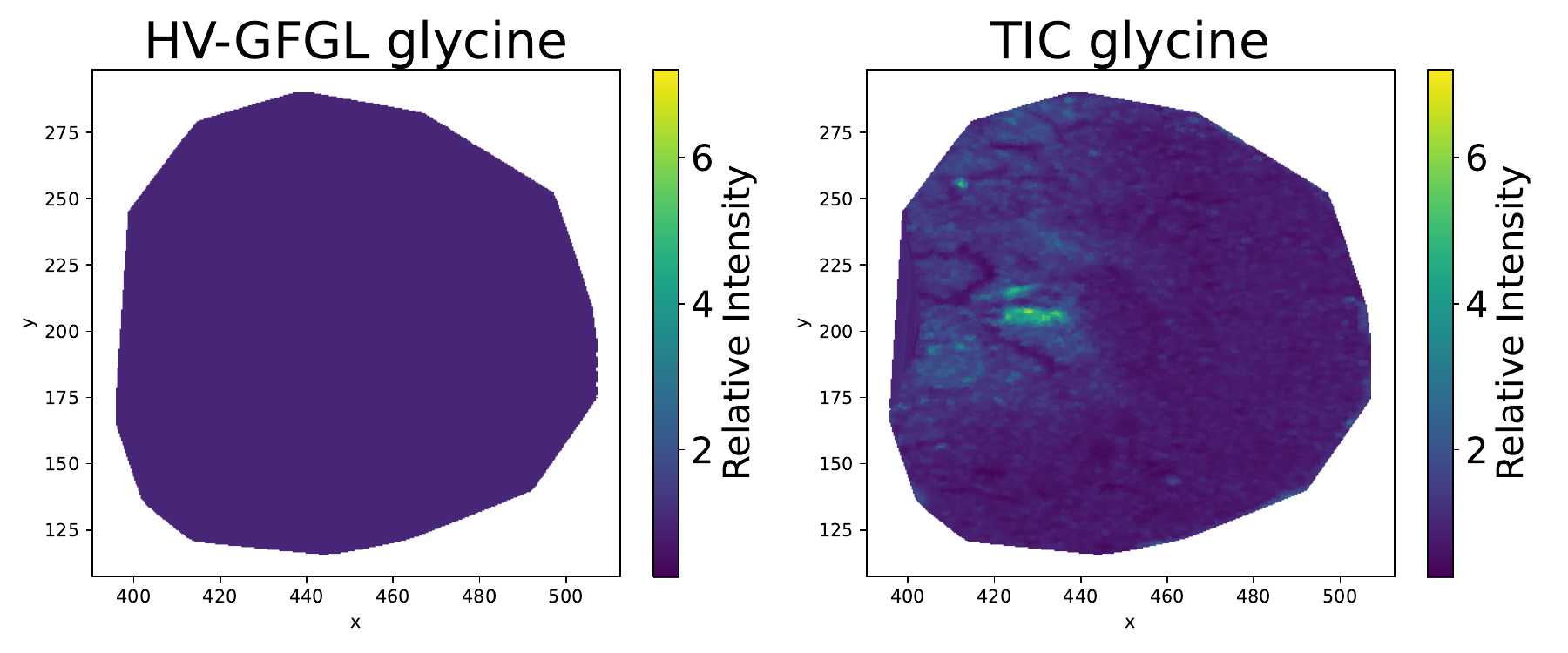}
        \caption{}
        \label{fig:glycine}
    \end{subfigure}

    \caption{HV-GFGL and TIC values plotted for Cysteine \textbf{(a)}, Glucose \textbf{(b)}, Glutamine \textbf{(c)}, and Glycine \textbf{(d)}. Color bar represents unitless relative rate. For (c) and (d), the colorbar is standardized.}
    \label{fig:kidney_results}
\end{figure}

\subsection{Results and analysis}
We noted significant differences in metabolite levels between the TIC-normalized data and HV-GFGL estimates. These differences largely fell into two categories. In the first, we identified many metabolites, such as the glutathione precursor cysteine (\cref{fig:kidney_results}a), where HV-GFGL estimates more accurately mirrored known physiology relative to TIC measurements~\cite{wang:etal:2010:specific-metabolic-rates}. This was most apparent with the key sugar glucose, which in raw TIC-normalized MALDI images showed only minor enhancement in the kidney medulla, but showed marked enhancement and high levels in the medulla in the HV-GFGL estimates (\cref{fig:kidney_results}b). In the second category, we noted that HV-GFGL identified metabolites which artifactually demonstrated patterns in TIC-normalized data due to their low abundance. This included key anabolic precusors such as NAD+ as well as amino acids such as glycine (\cref{fig:kidney_results}c-d).~\looseness=-1

We computed Structural Similarity Index Measure \citep{zWang} between TIC relative rates and HV-GFGL, with a majority of metabolites having an SSIM between 0.5-0.7 (\cref{tab:ssim_quantiles}). This indicates substantial structural divergence between TIC-normalized data and HV-GFGL estimates in most metabolites. Thus, we expect most metabolic analyses would likely be meaningfully different using HV-GFGL estimates.~\looseness=-1
\begin{table}[ht]
\centering
\caption{Quantiles of SSIM scores between the TIC and HV-GFGL.}

\begin{tabular}{lccccc}
\toprule
Quantile & 10\% & 25\% & 50\% (Median) & 75\% & 90\% \\
\midrule
SSIM     & 0.445 & 0.508 & 0.553 & 0.694 & 0.825 \\
\bottomrule
\end{tabular}
\label{tab:ssim_quantiles}
\end{table}

\section{Discussion}

\subsection{Overview of contributions}
We introduced a methodology for recovering relative rates of compositional data across a spatial field, with a specific focus on biological imaging data. Our method demonstrated clear superiority over current standards in the field in recovering relative rates, with implications for a wide range of scientific settings. We achieved this in the context of missing data with an original approach to a multinomial likelihood with a partially observed total. We developed a novel sparse structured variational family which significantly improves posterior coverage compared to mean-field VI.

\subsection{Limitations and future work}
Our approach is designed to capture piecewise constant, short length scale changes that are characteristic of biological imaging data. While a sparse shrinkage prior is a reasonable assumption in this context, this approach is less suited to spatial patterns that are either smoother or at longer length-scales, as is common in many spatial applications. To that end, adapting the prior to have a smoother penalty could prove fruitful in different applications, though whether the relative rates would still be identifiable is unclear. Additionally, while our model scales linearly with input size in all dimensions, computational limitations arise from the compositional likelihood and structured variational distribution, which introduce GPU synchronization bottlenecks. Addressing these limitations for extremely large datasets will likely require further optimization. 

We intend to expand our sparse VI framework to spatial problems outside of compositional data. In particular, the scalability of this method to further dimensions, such as spatio-temporal or spatial cohort data, could be an exciting development in variational methods for large-scale spatial data in a variety of application areas. Further, our entire approach would benefit  from strong theoretical guarantees to precisely delineate the requirements for identifiability and, ideally, finite-sample rates. This could include both a theoretical exploration of relative rate recovery for the model, and a comparative analysis of posterior approximation to MCMC methods. This latter study would also benefit from comparisons to other structured and hierarchical VI approaches.

\newpage
\bibliography{GFL_Metabolites/References}
\bibliographystyle{unsrt}

\newpage
\end{document}